\documentclass[11pt,letterpaper]{llncs}
\usepackage{amsmath}
\usepackage{amssymb}
\usepackage{graphicx}
\usepackage{marvosym}
\usepackage{url}
\usepackage{fancyhdr}
\usepackage{placeins}
\usepackage{cite}

\usepackage{geometry}
\newcommand{\keywords}[1]{\par\addvspace\baselineskip
\noindent\keywordname\enspace\ignorespaces#1}

\fancypagestyle{firstpage}{\fancyhf{}
\setcounter{page}{1}
\rfoot{\thepage}
}

\begin{document}

%\mainmatter  % start of an individual contribution

\title{\LARGE{Inclusivity in Large Language Models: Personality Traits and Gender Bias in Scientific Abstracts}}

%\titlerunning{Lecture Notes in Computer Science: Authors' Instructions}


\author{Naseela Pervez$^1$ \and Alexander J. Titus$^{1,2,3}$ \thanks{Corresponding Author: publications@theinvivogroup.com}}
\institute{\large{Information Sciences Institute, University of Southern California \and Iovine and Young Academy, University of Southern California \and In Vivo Group}}

%
% \authorrunning{Lecture Notes in Computer Science: Authors' Instructions}

%\institute{Springer-Verlag, Computer Science Editorial,\\
%Tiergartenstr. 17, 69121 Heidelberg, Germany\\
%\url{http://www.springer.com/lncs}}


%\toctitle{Lecture Notes in Computer Science}
%\tocauthor{Authors' Instructions}

\maketitle

\thispagestyle{firstpage}

\begin{abstract}

Large language models (LLMs) are increasingly utilized to assist in scientific and academic writing, helping authors enhance the coherence of their articles. Previous studies have highlighted stereotypes and biases present in LLM outputs, emphasizing the need to evaluate these models for their alignment with human narrative styles and potential gender biases. In this study, we assess the alignment of three prominent LLMs—Claude 3 Opus, Mistral AI Large, and Gemini 1.5 Flash—by analyzing their performance on benchmark text-generation tasks for scientific abstracts. We employ the Linguistic Inquiry and Word Count (LIWC) framework to extract lexical, psychological, and social features from the generated texts. Our findings indicate that, while these models generally produce text closely resembling human-authored content, variations in stylistic features suggest significant gender biases. This research highlights the importance of developing LLMs that maintain a diversity of writing styles to promote inclusivity in academic discourse.

\end{abstract}

\section{Introduction}
\label{section:intro}
Large Language Models (LLMs) have gained popularity in recent years for their performance on benchmark natural language processing (NLP) tasks including, but not limited to, text summarization, text generation, and question answering. LLMs have the ability to generate texts (stories) that are grammatically correct, and coherent and keep the readers engaged. It is likely that in the near future LLMs will be adopted as assistants for many writing tasks such as rewriting and improving the language of text (e.g. student essays, newspaper article, book writing). Scientific texts like abstracts, proposals, and journal publications use scientific jargon. However, due to the technical nature of scientific publications, to our knowledge, there are no full-text scientific articles that have been written by LLMs yet. In this paper we have proposed an evaluation framework that reflects the traits of LLM written text for scientific literature. 

Although scientific papers are aimed at conveying a technique, research solution, or a research proposal, every author has a writing style. This writing style has an impact on their readership. In theoretical psychology, there are well-established frameworks that quantify a piece of writing by various features. These features are what can be called the "personality" of the text. 

Like all texts, scientific texts have a personality and scientific authorship is male-dominated \cite{Rinaldo2023-uo}. It has been established that male authors have a higher readership and citation count as well \cite{Van_den_Besselaar2015-hc}. There are two aspects to this bias in research communities \cite{Kerkhoven2016-fh}  - 1. the established stereotype that males are better researchers than females, and 2. the writing style of males (informative) is appreciated over the writing style of female researchers (descriptive) \cite{Arkin,key}. 

Research communities have made efforts to reduce the gender gaps between males and females. This is reflected in the appointment of female faculty, promotion of male-female collaborations as well as high-prestige institutions hiring female researchers as project leads that have improved the representation of females in academia and improved the reach of female research \cite{Ceci2009-iv}. It is therefore crucial to ensure that LLMs used for scientific writing do not create, or exacerbate, a gender bias based on the writing style. 

\begin{figure}
    \centering
    \includegraphics[width=\textwidth]{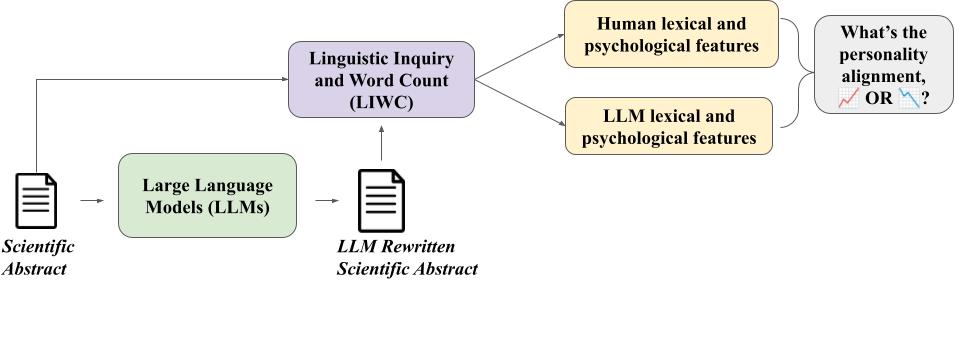}
    \caption{Flowchart illustrating the comparison of LIWC features in scientific abstracts written by humans and rewritten by LLMs to assess personality alignment. This framework is adapted for male vs female comparison as well.}
    \label{fig:framework}
\end{figure}

In this paper, we have focused on the following research questions:
\begin{enumerate}
    \item When prompted to re-write a piece of scientific text, do LLMs maintain the narrative style of the text, implying that it maintains the author's personality?
    \item Do LLMs alleviate or undermine the personality traits of a scientific text? Do they accentuate positive traits and diminish negative traits?
\end{enumerate}

\section{Related Work}
\label{section:related work}

 The manner in which a narrative is constructed can significantly influence the impact and dissemination of scientific literature, with varying styles potentially affecting the learning process of readers \cite{Hillier2016-nc,Bower1969-da}. Gender differences in writing styles have been observed, with female authors often adopting a more intimate and engaged tone, while male authors tend to employ a more directive and informational approach \cite{levitskaya2022investigating}. These stylistic disparities can inadvertently contribute to gender and prestige biases within the research community, potentially impacting the visibility and reception of similar research findings \cite{Helmer2017,Lariviere2013-dw}. However, it is crucial to prioritize the value and outcomes of research above stylistic preferences, fostering an inclusive appreciation for the diversity of writing styles in academic and scientific discourse.

In the realm of computational linguistics, the analysis of language goes beyond mere comprehension, delving into the intricate features that underpin it. This quantification of language has found diverse applications, including the identification of authors' genders, detection of underlying psychological issues, and recognition of hate speech, among others \cite{genderdiff, Fraser2016-kc,Englmeier2020TheRO}. Established frameworks in linguistics and psychology \cite{liwc,seance,empath} have been instrumental in text analysis, providing tools to examine the narrative style of a text from a lexical, psychological and social perspective. Research in this field posits that the these characteristics of a text serve as a mirror to the author's thought processes and personality traits \cite{Llerena2012-lk,language}. Consequently, a comprehensive understanding of an author's personality can be gleaned from a thorough analysis of their writings. This approach offers a panoramic view of the author's persona, further enriching the field of text analysis.

Previous studies have demonstrated that language styles can be associated with the author's gender \cite{Haas1979-qj,malevsfemale}. Research has extensively examined the directive style is typically used by males and the involved style is often employed by females \cite{Wen2013AGA}. Quantifying the narrative style of a text provides a framework for exploring gender differences in language use within STEM fields and the research community. Although research output should be valued regardless of writing style, directive writing tends to receive more citations due to its conciseness, which allows readers to grasp the content without extensive analysis \cite{Ma2023-go}.

Large language models (LLMs) are extensively utilized in text generation tasks, as discussed in Section \ref{section:intro}. There have been instances where LLMs reflect stereotypes, leading to gender bias in various societal contexts \cite{llm_stereotype}. However, studies have shown that with the use of efficient prompts, the personality traits of LLMs, such as extroversion and neuroticism, can be effectively tuned \cite{mao2024editing}. Understanding the linguistic markers of LLM-generated text which reflect the lexical and psychological personality traits is an active and ongoing area of research.

To understand the extent to which LLMs induce gender bias in generated text, we conducted a correlation analysis comparing lexical, psychological, and social features 
 computed by LIWC \cite{liwc} of scientific abstracts re-generated by LLMs with those of human-written abstracts. Our study further includes an examination of differences between human-written and LLM-generated scientific abstracts to identify and quantify the gender gaps in these features.

\section{Methodology}

\label{section: methodology}
In this section, we present the methodology and framework of our research (Figure \ref{fig:framework}). We provide a detailed discussion of the data used for analysis, the large language models (LLMs) employed for text regeneration, and the prompts provided to these LLMs. Additionally, we describe the framework utilized to compute the lexical and psychological traits of the text. Finally, we explain our approach to quantifying the alignment between human and LLM-generated text features. 
\begin{table}[ht]
\caption{Distribution of Genders in Scientific Abstracts from the CORE Dataset}
\label{tab: genders}
\centering
\begin{tabular}{cc}
\hline
\textbf{Gender} & \multicolumn{1}{r}{\textbf{Count}} \\
\hline
Female &  418\\
Male &  946\\
Mixed-Gender &  2026\\
\hline
\end{tabular}
\end{table}

\subsection{Data}
For the analysis presented in this paper, we focused exclusively on scientific abstracts rather than full-text articles. We selected a subset of 3,390 abstracts from the CORE dataset \cite{Knoth2023-zi}. Although the dataset includes author details, it does not provide gender information. To address this, we used the Python library, \textit{gender-extractor}, to assign genders to the authors\footnote{https://pypi.org/project/gender-extractor/}. Table \ref{tab: genders} shows the distribution of publications among male-only, female-only, and mixed-gender authors.

Given that many publications have both male and female authors, for the analysis of male vs. female personality alignment, we considered only publications authored solely by males or solely by females (n=1,364). However, to evaluate the overall alignment between human and LLM-generated abstracts, we utilized the entire dataset.

\subsection{Large Language Models}

The use of large language models (LLMs) has become increasingly prevalent in various natural language processing tasks. In this paper, we are using LLMs to rewrite scientific abstracts of authors. For regenerating the text of human-written scientific abstracts, we utilized three LLMs: Claude 3 Opus\footnote{https://www.anthropic.com/news/claude-3-family}, Mistral AI Large\cite{jiang2023mistral}, and Gemini 1.5 Flash\cite{geminiteam2024gemini}. These models were selected due to their popularity and high performance on benchmark text-generation tasks including but not limited to question answering, mathematical reasoning, diagram understanding. Their performance on these benchmarks not only surpasses traditional machine learning algorithms but also stands on par with each other, showcasing their advanced capabilities in generating coherent and contextually accurate text.

The following prompt was consistently employed across all three LLMs for the regeneration of scientific abstracts:

"\textbf{Given the scientific abstract, imagine yourself to be an author and researcher, and rewrite this abstract.
The abstract is : \textit{[content of the abstract]}}"

\subsection{Linguistic Inquiry and Word Count (LIWC)}
Linguistic Inquiry and Word Count LIWC\cite{liwc} is a text analysis framework comprising thousands of dictionaries. The software quantifies the lexical, psychological, and social features of a text based on these dictionaries. In this study, we utilized the LIWC-22\cite{LIWC-22} dictionary for research and analysis. Given our focus on scientific writing, we employed LIWC features pertinent to this domain; for instance, the \textit{curse} feature is not relevant for quantifying scientific abstracts.

For our analysis, we selected features relevant to the narration of scientific texts, concentrating on narrow rather than broad categories of LIWC features. The specific features we focused on for this research are: Segment, WC, Analytic, Clout, Tone, affiliation, achieve, power, insight, cause, discrep, tentat, certitude, differ, tone\_pos,tone\_neg, emotion,	emo\_pos, emo\_neg, emo\_anx, emo\_anger, emo\_sad, prosocial, polite, conflict, moral, comm, politic, ethnicity, tech, reward, risk, curiosity, allure.

\subsection{Comparison Between Human-Written and LLM-Generated Scientific Abstracts}

Our objective is to determine the alignment between human and LLM-generated texts by comparing various features. We conduct two primary types of analyses:

\subsubsection{Correlation Analysis:}

We employ the Pearson correlation coefficient to assess the relationship between human-written abstracts and those regenerated by three LLMs: Claude 3 Opus, Gemini 1.5 Flash, and Mistral AI Large. Specifically, we compute the correlations for each LLM to determine how closely their generated texts align with human-authored texts. This correlation analysis allows us to evaluate the association of lexical, psychological, and social features between human and LLM-generated abstracts. By doing so, we can assess the extent to which these LLMs replicate human personality traits in their generated scientific abstracts.

\subsubsection{T-Test Analysis:}

We use t-tests to find the differences in means for the features of the following scientific abstracts:
\begin{itemize}
\item Female-authored abstracts vs. male-authored abstracts (Human Female vs. Human Male).
\item LLM-generated abstracts for female authors vs. LLM-generated abstracts for male authors (AI Female vs. AI Male).
\end{itemize}

These analyses are conducted for all three LLMs. The results are presented in the following section.

\section{Results}
\begin{table}[htbp]
\caption{Pearson correlation for humans vs LLMs} 
\centering
    
\begin{tabular}{lrrr}
\hline
LIWC & Claude & Gemini & Mistral \\
\hline
Segment & NaN & NaN & NaN \\
WC & 0.35 & 0.80 & 0.86 \\
Analytic & 0.33 & 0.49 & 0.37 \\
Clout & 0.73 & 0.80 & 0.77 \\
Tone & 0.75 & 0.75 & 0.72 \\
affiliation & 0.60 & 0.73 & 0.71 \\
achieve & 0.65 & 0.65 & 0.66 \\
power & 0.81 & 0.81 & 0.80 \\
insight & 0.81 & 0.81 & 0.82 \\
cause & 0.70 & 0.75 & 0.70 \\
discrep & 0.22 & 0.22 & 0.22 \\
tentat & 0.58 & 0.71 & 0.60 \\
certitude & 0.51 & 0.43 & 0.48 \\
differ & 0.66 & 0.75 & 0.73 \\
tone\_pos & 0.68 & 0.69 & 0.66 \\
tone\_neg & 0.80 & 0.80 & 0.76 \\
emotion & 0.51 & 0.55 & 0.55 \\
emo\_pos & 0.48 & 0.56 & 0.54 \\
emo\_neg & 0.72 & 0.73 & 0.63 \\
emo\_anx & 0.86 & 0.88 & 0.88 \\
emo\_anger & 0.75 & 0.76 & 0.72 \\
emo\_sad & 0.53 & 0.52 & 0.31 \\
prosocial & 0.82 & 0.79 & 0.80 \\
polite & 0.64 & 0.64 & 0.63 \\
conflict & 0.82 & 0.79 & 0.78 \\
moral & 0.70 & 0.64 & 0.59 \\
comm & 0.65 & 0.61 & 0.66 \\
politic & 0.85 & 0.88 & 0.85 \\
ethnicity & 0.86 & 0.92 & 0.91 \\
tech & 0.86 & 0.91 & 0.89 \\
reward & 0.66 & 0.66 & 0.66 \\
risk & 0.85 & 0.85 & 0.84 \\
curiosity & 0.74 & 0.79 & 0.80 \\
allure & 0.63 & 0.62 & 0.62 \\
\hline
\end{tabular}

\label{tab:corr}

\end{table}

In this section, we detail how the lexical, psychological, and social features of LLM-rewritten scientific abstracts, obtained from LIWC, differ from those written by humans. We explore these differences in two key aspects: \textit{humans vs. LLMs}, which examines the alignment of LLM personality traits with human personality traits, and \textit{males vs. females}, which investigates whether the narrative style of LLMs reflects any gender bias.

\subsection{Humans vs LLMs: Correlation Analysis}

To assess the alignment of LIWC features between human-generated and LLM-generated scientific abstracts, we computed Pearson correlation coefficients across all pairs of features. Our focus, however, lies specifically on the diagonal elements of this correlation matrix.

\begin{figure}
    \centering
    \includegraphics[width=\textwidth]{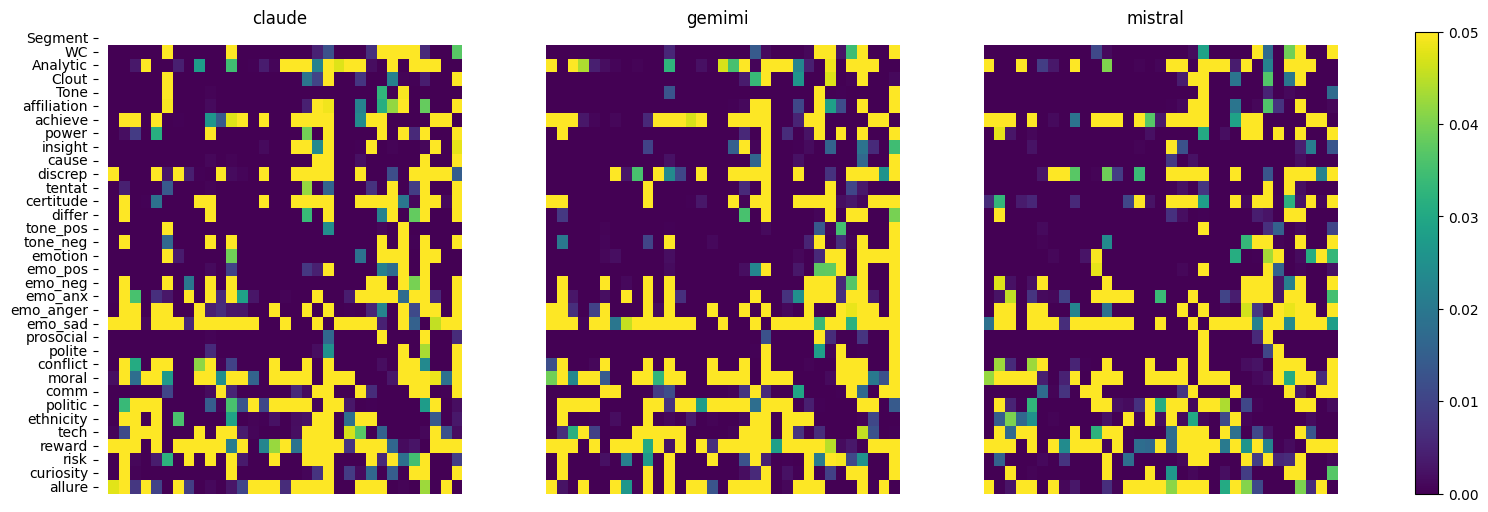}
    \caption{Heatmap representing the significance (p\_value) of pearson correlation coefficient of LIWC features between humans and LLMs - Claude, Gemini, Mistral (left to right)}
    \label{fig:pval}
\end{figure}

Figure \ref{fig:pval} displays the p-values of the Pearson correlation coefficients between pairs of LIWC features. The heatmap reveals that along the diagonal, all feature pairs have p-values less than 0.05. This indicates that the Pearson correlation coefficient for all diagonal elements (representing LIWC features of humans vs LLMs) across all LLMs is statistically significant (p\_value \(<\) 0.05).

A higher positive Pearson correlation coefficient between two features indicates that when one feature's value increases, the other feature tends to increase as well. In our study, interpreting the diagonal elements, a higher positive correlation suggests greater alignment between human and LLM-generated abstracts in terms of these features. Conversely, a higher negative correlation coefficient indicates an inverse relationship between the two datasets. In this context, a positive coefficient signifies similarity in the expression of features between humans and LLMs, reflecting alignment in personality traits. Conversely, a negative coefficient suggests divergence, implying contrasting expressions of these traits between humans and LLMs. 

\begin{figure}
    \centering
    \includegraphics[width=\textwidth]{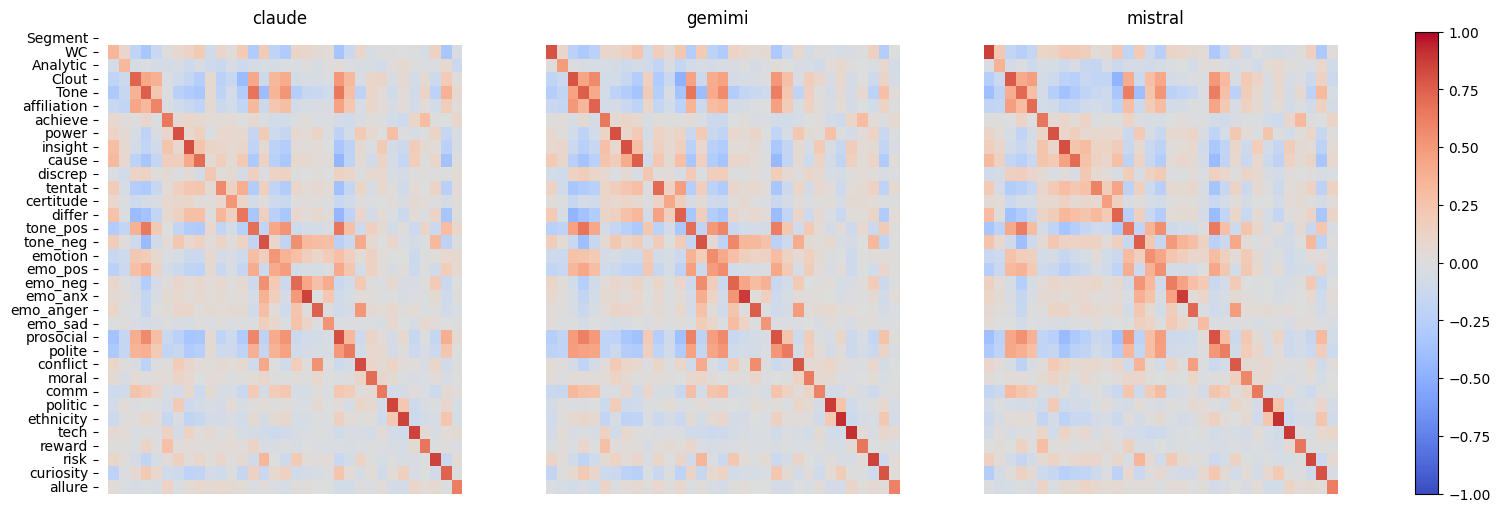}
    \caption{Heatmap representing the pearson correlation coefficient of LIWC features between humans and LLMs - Claude, Gemini, Mistral (left to right)}
    \label{fig:corr}
\end{figure}

In our correlation analysis of LIWC features comparing human-generated and LLM-generated text, we observed a significant positive correlation among the diagonal elements. However, we found minimal to no correlation between other pairs of features. Specifically, the diagonal elements prominently exhibit a strong positive correlation (see Figure \ref{fig:corr}). 

Table \ref{tab:corr} presents Pearson correlation coefficients for LIWC features comparing human-generated texts with those produced by all LLMs. It is important to note that all correlations are statistically significant (p-value < 0.05). The lexical feature \textit{WC} (word count) shows a weaker positive correlation (0.35) for Claude Opus compared to Gemini and Mistral (refer to Table \ref{tab:corr}).

Across cognitive features in Table \ref{tab:corr}, there is a generally higher positive correlation, indicating alignment between LLMs and humans. Notably, the feature "certitude," reflecting confidence in text, particularly lacks strength in Gemini, suggesting that scientific abstracts generated by LLMs may not convey certitude similarly to human-authored text.

Regarding tone features, \textit{tone\_pos} and \textit{tone\_neg} exhibit strong positive correlations between humans and all three LLMs. However, features related to emotions relevant to scientific writing (\textit{emotion} and \textit{emo\_pos}) show positive correlations but not as pronounced.

Analysis of social process features such as 'prosocial', 'polite', 'conflict', 'moral', and 'comm' reveals consistently high positive correlation coefficients across all three LLMs. Claude Opus particularly stands out with the highest values, indicating its texts closely mirror human social processes.

Features reflecting motives in LLM-generated scientific abstracts—'reward', 'risk', 'curiosity', and 'allure'—also demonstrate high positive correlations with human-written abstracts, suggesting precise capture and reflection of textual motives by LLMs.

Lastly, general topic categories—'politic', 'ethnicity', and 'tech'—exhibit higher correlations between humans and LLMs, indicating accurate reflection of document categories in regenerated texts.

Our findings demonstrate a strong positive correlation between LIWC features in human-written and LLM-generated scientific abstracts across all three LLM models. This suggests that LLMs effectively capture the lexical characteristics, psychological traits, and social dynamics observed in human-authored texts.

\begin{table}[htbp]
\caption{t-test statistics for Males vs Females}
\centering
\begin{tabular}{lrrrr}
\hline

LIWC  & Human & Claude & Gemini & Mistral\\
\hline
Segment & NaN & NaN & NaN & NaN\\
WC & \textbf{\textit{-5.86}} & \textbf{\textit{-4.66}} & \textbf{\textit{-6.13}}  & \textbf{\textit{-6.31}}\\
Analytic & -1.13 & -1.63 & -1.74 & -1.22\\
Clout & -0.71 & -1.86 & -1.69 & -1.44\\
Tone & \textbf{\textit{3.93}} & 1.77 & \textbf{\textit{2.37}} & \textbf{\textit{2.43}}\\
affiliation & 0.19 & -0.71 & -0.57 & 1.57\\
achieve & -0.78 & \textbf{\textit{-3.22}} & \textbf{\textit{-2.15}} & \textbf{\textit{-2.67}}\\
power & -0.42 & 0.75 & -0.51 & -0.27\\
insight & \textbf{\textit{-6.74}} & \textbf{\textit{-6.66}} & \textbf{\textit{-7.64}} & \textbf{\textit{-6.98}}\\
cause & \textbf{\textit{-2.27}} & -0.51  & -1.56 & -1.38\\
discrep & -0.35 & -0.33 & \textbf{\textit{-3.01}} & 0.73\\
tentat & -0.31 & 0.23 & 1.67 & 0.86 \\
certitude & -0.01 & -0.54 & -1.66 & -1.90\\
differ & \textbf{\textit{-3.20}} & \textbf{\textit{-2.17}} & \textbf{\textit{-3.11}} & -1.59\\
tone\_pos & \textbf{\textit{3.88}} & \textbf{\textit{2.21}}& \textbf{\textit{2.04}} & \textbf{\textit{2.01}} \\
tone\_neg & -1.25 & -1.12 & -0.77 & -0.7 \\
emotion & \textbf{\textit{3.33}} & \textbf{\textit{5.8}} & \textbf{\textit{3.73}} & \textbf{\textit{5.05}}\\
emo\_pos & \textbf{\textit{3.82}} & \textbf{\textit{6.20}} & \textbf{\textit{4.71}}  & \textbf{\textit{5.64}}\\
emo\_neg & 0.96 & 1.30 & 1.40 & 1.01\\
emo\_anx & -0.08 & -0.01 & 0.22 & 0.12\\
emo\_anger & -0.31 & 0.27 & 0.36 & -0.61\\
emo\_sad & 1.07 & 0.27 & 0.67 & 0.79\\
prosocial & \textbf{\textit{3.15}} & \textbf{\textit{3.00}} & \textbf{\textit{2.96}} & 1.57 \\
polite & \textbf{\textit{5.70}} & \textbf{\textit{3.70}}& \textbf{\textit{4.10}} & \textbf{\textit{2.82}}\\
conflict & -1.95 & \textbf{\textit{-2.53}} & \textbf{\textit{-2.58}} & \textbf{\textit{-2.00}}\\
moral & -1.63 & -0.92 & 0.11 & -0.14\\
comm & 1.08 & -1.42 & -0.84 & -1.52\\
politic & -0.60 & -0.85& -1.26 & -1.45 \\
ethnicity & 0.37 & -0.51 & 0.28 & -0.13\\
tech & 1.66 & 0.85 & 0.93 & 1.54 \\
reward & -1.84 & -1.73 & -.12 & -1.46\\
risk & \textbf{\textit{2.61}} & \textbf{\textit{2.10}} & 1.94 & 1.93\\
curiosity & 1.06 & \textbf{\textit{3.75}} & \textbf{\textit{2.60}} & \textbf{\textit{2.66}}\\
allure & -1.77 & -0.48 & -0.9 & -0.3\\
\hline
\end{tabular}
\label{tab:t-test}

\end{table}

\subsection{Gender Bias: Two Sample t-test }

We conducted a two-sample t-test to compare the LIWC features between male and female authors. Among the 35 features analyzed, 15 features showed a statistically significant t-statistic value (p-value 
\(<\) 0.05). Table \ref{tab:t-test} presents the t-statistic values for scientific abstracts written by humans, as well as those generated by Claude, Gemini, and Mistral. In the table, the significant values are highlighted in bold and italicized for clarity.
\begin{figure}
    \centering
    \includegraphics[width=\textwidth]{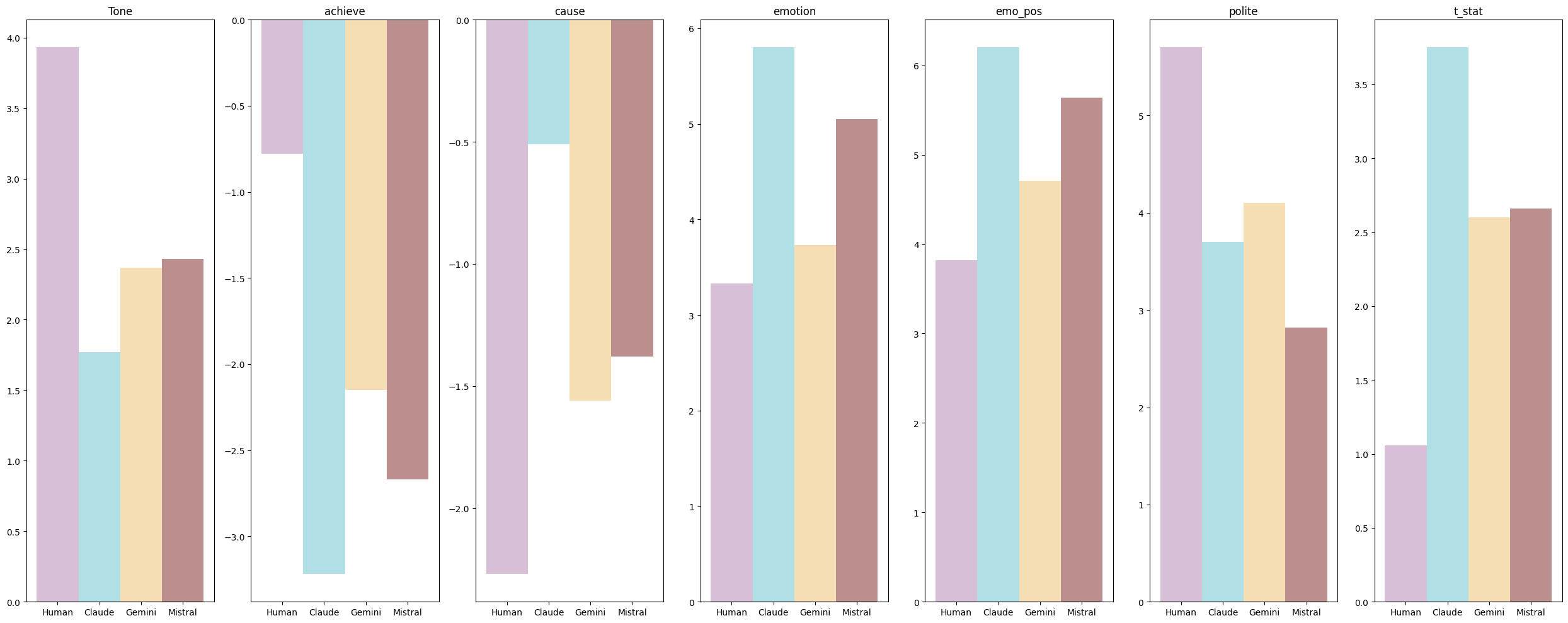}
    \caption{t-statistic values of statistically significant features ('Tone', 'achieve', 'cause', 'emotion', 'emo\_pos','polite','curiosity') which reflects gender gaps between human and LLM texts.}
    \label{fig:t-test}
\end{figure}

For the lexical features, both \textit{WC} (word count) and \textit{tone} are statistically significant. On average, female authors use 5.86 times more words than male authors, a pattern that is also observed in the LLM-generated texts, as shown in table \ref{tab:t-test}. Additionally, male authors tend to have a more positive tone in their abstracts compared to female authors, approximately 4 times more. Although LLMs reflect this same trend, it is noteworthy that the difference is less pronounced, especially for Claude. This highlights an important observation: while LLMs generally follow the authors' personality traits in lexical features, they can sometimes underestimate or overestimate these features.

Among the cognitive features, \textit{insight}, \textit{cause}, and \textit{differ} are statistically significant. The scientific abstracts show that female authors use approximately seven times more insightful words such as "know," "think," and "feel" compared to male authors. This pattern is also reflected in the LLM-generated texts, with similar values (table \ref{tab:t-test}). For the feature \textit{differ}, the difference in writing style between males and females is consistent across all LLMs except Mistral, where the t-statistic is not significant. Interestingly, for the feature \textit{cause} (causation), the difference between male and female narration is significant, with females using more causation-centric terms. However, this trend is not followed by any of the LLMs. Although not a drastic difference, but LLMs do introduce bias in terms of using causation-centric words that is found approximately twice more in females than males.

We found that the psychological process of affect is significantly represented by three features: \textit{tone\_pos}, \textit{emotion}, and \textit{emo\_pos} (p \(<\) 0.05). These features, out of the eight selected, were the only ones to show statistical significance. They reflect the positive tone and emotional content present in the text. Interestingly, both human authors and large language models (LLMs) consistently indicated that texts authored by males exhibited at least three times more positivity than those authored by females (Table 1). However, it is noteworthy that the LLMs, specifically Claude and Mistral, tended to overestimate the emotional content and positivity in texts authored by males. This overestimation could potentially amplify the perceived gender disparity between male and female authors in the field.  

In the domain of social behavior processes, the categories \textit{prosocial}, \textit{polite}, and \textit{conflict} are statistically significant. Notably, while the difference in narration style between males and females regarding conflict is non-significant, it becomes significant in the texts generated by large language models (LLMs). This suggests that female authors use a more conflicting style of writing compared to their male counterparts. Additionally, male authors tend to adopt a significantly more polite writing style, approximately five times more polite on average than females that is consistent with existing literature \cite{Ma2023-go}. However, this distinction is underestimated in LLM-generated abstracts, contributing to a gender gap in the portrayal of politeness in writing styles (see table \ref{tab:t-test}).

Moving on to the motive category, we identified \textit{risk} and \textit{curiosity} as two significant features. Our analysis revealed that male authors used approximately twice as many risk-associated words compared to female authors. However, this difference was only statistically significant (p \(<\) 0.05) in texts regenerated by Claude, and not in those regenerated by Gemini or Mistral. In addition to this, the analysis shows a significant (p\_value \(<\) 0.05) difference between male and female re-written abstracts by LLMs for \textit{achieve} category reflecting that females use approximately thrice more words reflecting achievement which is not reflected in the human authored articles.

Additionally, our analysis found no significant difference (p \(>\) 0.05) between male and female authors in terms of reflecting curiosity in their narrative style. Nevertheless, LLMs indicated that male authors adopted a narrative style that reflected more curiosity than their female counterparts (Table 2). This finding highlights the potential for LLMs to further exacerbate the gender gap in scientific writing by perpetuating biased representations of author characteristics.

We did not observe a gender gap in the lexical features and cognitive processes of scientific abstracts when comparing human authors to LLM-generated texts. However, for the psychological processes of affect and motive, LLMs tend to amplify the gender gap observed between males and females for certain features. Additionally, while not drastic, a gender gap is also present in the politeness feature within the social behavior category. The t-statistics for these features indicate a significant difference (see Figure \ref{fig:t-test}).

\section{Conclusion and Future Works}

The study analyzes the personality traits reflected by LLMs in the context of employing them for scientific writing. The analysis reveals that the lexical, psychological, and social traits of text generated by LLMs closely align with those of human-written text. Therefore, it can be concluded that LLMs have the capacity to depict traits similar to those of human authors and researchers. This is notable because there is a fierce debate in the academic and scientific community about the value of LLM-generated text.

The study analyzes the personality traits reflected by LLMs in the context of employing them for scientific writing. The analysis reveals that the lexical, psychological, and social traits of text generated by LLMs closely align with those of human-written text, indicating that LLMs can depict traits similar to those of human authors and researchers. Additionally, we discovered that male and female authors exhibit distinct differences in narrative styles, particularly in terms of politeness, conflict, and risk-related words. LLMs not only reflect but also magnify these differences, potentially perpetuating gender stereotypes in scientific writing. This highlights a critical area for improvement in LLM training and deployment to ensure fair and unbiased text generation. Future research should focus on refining LLM algorithms to minimize bias and enhance their ability to generate equitable representations across different genders.

Future work should focus on not only rewriting the abstracts but also generating the abstracts for the male vs female-authored full-text articles and analyzing the gender bias in this case. Additionally, studying gender bias for various academic disciplines and across time is also an important step in detecting and mitigating bias in research communities. This will contribute to identifying the pros and cons of AI resources and how they can be improved for academic and scientific disciplines.

% \vspace{2cm}

\section*{Authors}
\noindent {\bf Naseela Pervez} received her M.S Computer Science from Viterbi School of Engineering, University of Southern California (USC). Currently, she is a pre-doctoral research staff at Management of Innovation, Entrepreneurial Research, and Venture Analysis (MINERVA) and Information Sciences Institute (ISI) at the University of Southern California (USC). Her research interests include natural language processing, network science, social sciences, and fairness and bias in AI.\\

\noindent {\bf  Alexander J. Titus} received his Ph.D. in Quantitative Biomedical Sciences from Dartmouth College. Currently, he is a Principal Scientist at the Information Sciences Institute and Research Faculty at the Iovine and Young Academy at the University of Southern California (USC), and Founder and Principal Investigator at the In Vivo Group. His research interests include the applications of artificial intelligence to the life sciences and responsible AI development.\\

% \noindent {\bf Priyadarshi Singh} received M.Tech from IIT(ISM) Dhanbad. He did Bachelor degree in Information Technology. Currently, he is pursuing his PhD in Computer Science from the University of Hyderabad. His research interests include Cryptography, Public key infrastructure.\\

% \noindent {\bf V. Ch. Venkaiah} obtained his PhD in 1988 from the Indian Institute of Science (IISc), Bangalore in the area of scientific computing. He worked for several
% organisations including the Central Research Laboratory of Bharat Electronics, Tata Elxsi India Pvt. Ltd., Motorola India Electronics Limited, all in Bangalore.
% He then moved onto academics and served IIT, Delhi, IIIT, Hyderabad, and C R Rao Advanced Institute of Mathematics, Statistics, and Computer Science. He is currently serving the Hyderabad Central University. He is a vivid researcher. He designed algorithms for linear
% programming, subspace rotation and direction of arrival estimation, graph colouring, matrix symmetriser, integer factorisation, cryptography, knapsack problem, etc.\\

% \noindent {\bf Subba Rao Y V} obtained his PhD from the University of Hyderabad. Currently, he is an Assistant Professor in the School of Computer and Information Sciences,
% University of Hyderabad. His area of interests includes Cryptography, Theory of Computation, etc.

\bibliography{custom}

\begin{thebibliography}{10}

\bibitem{Rinaldo2023-uo}
N.~Rinaldo, G.~Piva, S.~Ryder, A.~Crepaldi, A.~Pasini, L.~Caruso, R.~Manfredini, S.~Straudi, F.~Manfredini, and N.~Lamberti, ``The issue of gender bias represented in authorship in the fields of exercise and rehabilitation: A 5-year research in indexed journals,'' {\em J. Funct. Morphol. Kinesiol.}, vol.~8, p.~18, Jan. 2023.

\bibitem{Van_den_Besselaar2015-hc}
P.~van~den Besselaar and U.~Sandstr{\"o}m, ``Gender differences in research performance and its impact on careers: a longitudinal case study,'' {\em Scientometrics}, vol.~106, pp.~143--162, Nov. 2015.

\bibitem{Kerkhoven2016-fh}
A.~H. Kerkhoven, P.~Russo, A.~M. Land-Zandstra, A.~Saxena, and F.~J. Rodenburg, ``Gender stereotypes in science education resources: A visual content analysis,'' {\em PLoS One}, vol.~11, p.~e0165037, Nov. 2016.

\bibitem{Arkin}
N.~Arkin, C.~Lai, L.~M. Kiwakyou, G.~M. Lochbaum, A.~Shafer, S.~K. Howard, E.~R. Mariano, and M.~Fassiotto, ``{What's in a Word? Qualitative and Quantitative Analysis of Leadership Language in Anesthesiology Resident Feedback},'' {\em Journal of Graduate Medical Education}, vol.~11, pp.~44--52, 02 2019.

\bibitem{key}
\url{https://debuk.wordpress.com/2016/03/06/do-women-and-men-write-differently/}.
\newblock [Accessed 21-06-2024].

\bibitem{Ceci2009-iv}
S.~J. Ceci, W.~M. Williams, and S.~M. Barnett, ``Women's underrepresentation in science: sociocultural and biological considerations,'' {\em Psychol. Bull.}, vol.~135, pp.~218--261, Mar. 2009.

\bibitem{Hillier2016-nc}
A.~Hillier, R.~P. Kelly, and T.~Klinger, ``Narrative style influences citation frequency in climate change science,'' {\em PLoS One}, vol.~11, p.~e0167983, Dec. 2016.

\bibitem{Bower1969-da}
G.~H. Bower and M.~C. Clark, ``Narrative stories as mediators for serial learning,'' {\em Psychonomic Science}, vol.~14, pp.~181--182, Apr. 1969.

\bibitem{levitskaya2022investigating}
E.~Levitskaya, K.~Kedrick, and R.~J. Funk, ``Investigating writing style as a contributor to gender gaps in science and technology,'' 2022.

\bibitem{Helmer2017}
M.~Helmer, M.~Schottdorf, A.~Neef, and D.~Battaglia, ``Research: Gender bias in scholarly peer review,'' {\em eLife}, vol.~6, p.~e21718, mar 2017.

\bibitem{Lariviere2013-dw}
V.~Larivi{\`e}re, C.~Ni, Y.~Gingras, B.~Cronin, and C.~R. Sugimoto, ``Bibliometrics: Global gender disparities in science,'' {\em Nature}, vol.~504, pp.~211--213, Dec. 2013.

\bibitem{genderdiff}
V.~P{\'e}rez-Rosas and R.~Mihalcea, ``Gender differences in deceivers writing style,'' in {\em Human-Inspired Computing and Its Applications} (A.~Gelbukh, F.~C. Espinoza, and S.~N. Galicia-Haro, eds.), (Cham), pp.~163--174, Springer International Publishing, 2014.

\bibitem{Fraser2016-kc}
K.~C. Fraser, J.~A. Meltzer, and F.~Rudzicz, ``Linguistic features identify alzheimer's disease in narrative speech,'' {\em Journal of Alzheimer's Disease}, vol.~49, pp.~407--422, 2016.

\bibitem{Englmeier2020TheRO}
K.~Englmeier, ``The role of storylines in hate speech detection (short paper),'' in {\em Machine Learning for Trend and Weak Signal Detection in Social Networks and Social Media}, 2020.

\bibitem{liwc}
Y.~R. Tausczik and J.~W. Pennebaker, ``The psychological meaning of words: Liwc and computerized text analysis methods,'' {\em Journal of Language and Social Psychology}, vol.~29, pp.~24 -- 54, 2010.

\bibitem{seance}
S.~A. Crossley, K.~Kyle, and D.~S. McNamara, ``Sentiment analysis and social cognition engine ({SEANCE)}: An automatic tool for sentiment, social cognition, and social-order analysis,'' {\em Behavior Research Methods}, vol.~49, pp.~803--821, June 2017.

\bibitem{empath}
E.~Fast, B.~Chen, and M.~S. Bernstein, ``Empath: Understanding topic signals in large-scale text,'' in {\em Proceedings of the 2016 CHI Conference on Human Factors in Computing Systems}, CHI '16, (New York, NY, USA), p.~4647–4657, Association for Computing Machinery, 2016.

\bibitem{Llerena2012-lk}
K.~Llerena, S.~G. Park, S.~M. Couture, and J.~J. Blanchard, ``Social anhedonia and affiliation: examining behavior and subjective reactions within a social interaction,'' {\em Psychiatry Res}, vol.~200, pp.~679--686, Aug. 2012.

\bibitem{language}
K.~Brauer, R.~Sendatzki, and R.~T. Proyer, ``Testing associations between language use in descriptions of playfulness and age, gender, and self-reported playfulness in german-speaking adults,'' {\em Frontiers in Psychology}, vol.~13, 2022.

\bibitem{Haas1979-qj}
A.~Haas, ``Male and female spoken language differences: Stereotypes and evidence,'' {\em Psychological bulletin}, vol.~86, no.~3, pp.~616--626, 1979.

\bibitem{malevsfemale}
J.~N. Martin and R.~T. Craig, ``Selected linguistic sex differences during initial social interactions of same‐sex and mixed‐sex student dyads,'' {\em Western Journal of Speech Communication}, vol.~47, no.~1, pp.~16--28, 1983.

\bibitem{Wen2013AGA}
X.~Wen, P.~M. McCarthy, and A.~C. Strain, ``A gramulator analysis of gendered language in cable news reportage,'' in {\em The Florida AI Research Society}, 2013.

\bibitem{Ma2023-go}
Y.~Ma, Y.~Teng, Z.~Deng, L.~Liu, and Y.~Zhang, ``Does writing style affect gender differences in the research performance of articles?: An empirical study of {BERT-based} textual sentiment analysis,'' {\em Scientometrics}, vol.~128, pp.~2105--2143, Apr. 2023.

\bibitem{llm_stereotype}
H.~Kotek, R.~Dockum, and D.~Sun, ``Gender bias and stereotypes in large language models,'' in {\em Proceedings of The ACM Collective Intelligence Conference}, CI '23, (New York, NY, USA), p.~12–24, Association for Computing Machinery, 2023.

\bibitem{mao2024editing}
S.~Mao, X.~Wang, M.~Wang, Y.~Jiang, P.~Xie, F.~Huang, and N.~Zhang, ``Editing personality for large language models,'' 2024.

\bibitem{Knoth2023-zi}
P.~Knoth, D.~Herrmannova, M.~Cancellieri, L.~Anastasiou, N.~Pontika, S.~Pearce, B.~Gyawali, and D.~Pride, ``Core: A global aggregation service for open access papers,'' {\em Nature Scientific Data}, vol.~10, p.~366, June 2023.

\bibitem{jiang2023mistral}
A.~Q. Jiang, A.~Sablayrolles, A.~Mensch, C.~Bamford, D.~S. Chaplot, D.~de~las Casas, F.~Bressand, G.~Lengyel, G.~Lample, L.~Saulnier, L.~R. Lavaud, M.-A. Lachaux, P.~Stock, T.~L. Scao, T.~Lavril, T.~Wang, T.~Lacroix, and W.~E. Sayed, ``Mistral 7b,'' 2023.

\bibitem{geminiteam2024gemini}
G.~Team, P.~Georgiev, V.~I. Lei, R.~Burnell, L.~Bai, A.~Gulati, G.~Tanzer, D.~Vincent, Z.~Pan, S.~Wang, S.~Mariooryad, Y.~Ding, X.~Geng, F.~Alcober, R.~Frostig, M.~Omernick, L.~Walker, C.~Paduraru, C.~Sorokin, A.~Tacchetti, C.~Gaffney, S.~Daruki, O.~Sercinoglu, Z.~Gleicher, J.~Love, P.~Voigtlaender, R.~Jain, G.~Surita, K.~Mohamed, R.~Blevins, J.~Ahn, T.~Zhu, K.~Kawintiranon, O.~Firat, Y.~Gu, Y.~Zhang, M.~Rahtz, M.~Faruqui, N.~Clay, J.~Gilmer, J.~Co-Reyes, I.~Penchev, R.~Zhu, N.~Morioka, K.~Hui, K.~Haridasan, V.~Campos, M.~Mahdieh, M.~Guo, S.~Hassan, K.~Kilgour, A.~Vezer, H.-T. Cheng, R.~de~Liedekerke, S.~Goyal, P.~Barham, D.~Strouse, S.~Noury, J.~Adler, M.~Sundararajan, S.~Vikram, D.~Lepikhin, M.~Paganini, X.~Garcia, F.~Yang, D.~Valter, M.~Trebacz, K.~Vodrahalli, C.~Asawaroengchai, R.~Ring, N.~Kalb, L.~B. Soares, S.~Brahma, D.~Steiner, T.~Yu, F.~Mentzer, A.~He, L.~Gonzalez, B.~Xu, R.~L. Kaufman, L.~E. Shafey, J.~Oh, T.~Hennigan, G.~van~den Driessche, S.~Odoom, M.~Lucic, B.~Roelofs, S.~Lall, A.~Marathe,
  B.~Chan, S.~Ontanon, L.~He, D.~Teplyashin, J.~Lai, P.~Crone, B.~Damoc, L.~Ho, S.~Riedel, K.~Lenc, C.-K. Yeh, A.~Chowdhery, Y.~Xu, M.~Kazemi, E.~Amid, A.~Petrushkina, K.~Swersky, A.~Khodaei, G.~Chen, C.~Larkin, M.~Pinto, G.~Yan, A.~P. Badia, P.~Patil, S.~Hansen, D.~Orr, S.~M.~R. Arnold, J.~Grimstad, A.~Dai, S.~Douglas, R.~Sinha, V.~Yadav, X.~Chen, E.~Gribovskaya, J.~Austin, J.~Zhao, K.~Patel, P.~Komarek, S.~Austin, S.~Borgeaud, L.~Friso, A.~Goyal, B.~Caine, K.~Cao, D.-W. Chung, M.~Lamm, G.~Barth-Maron, T.~Kagohara, K.~Olszewska, M.~Chen, K.~Shivakumar, R.~Agarwal, H.~Godhia, R.~Rajwar, J.~Snaider, X.~Dotiwalla, Y.~Liu, A.~Barua, V.~Ungureanu, Y.~Zhang, B.-O. Batsaikhan, M.~Wirth, J.~Qin, I.~Danihelka, T.~Doshi, M.~Chadwick, J.~Chen, S.~Jain, Q.~Le, A.~Kar, M.~Gurumurthy, C.~Li, R.~Sang, F.~Liu, L.~Lamprou, R.~Munoz, N.~Lintz, H.~Mehta, H.~Howard, M.~Reynolds, L.~Aroyo, Q.~Wang, L.~Blanco, A.~Cassirer, J.~Griffith, D.~Das, S.~Lee, J.~Sygnowski, Z.~Fisher, J.~Besley, R.~Powell, Z.~Ahmed, D.~Paulus, D.~Reitter,
  Z.~Borsos, R.~Joshi, A.~Pope, S.~Hand, V.~Selo, V.~Jain, N.~Sethi, M.~Goel, T.~Makino, R.~May, Z.~Yang, J.~Schalkwyk, C.~Butterfield, A.~Hauth, A.~Goldin, W.~Hawkins, E.~Senter, S.~Brin, O.~Woodman, M.~Ritter, E.~Noland, M.~Giang, V.~Bolina, L.~Lee, T.~Blyth, I.~Mackinnon, M.~Reid, O.~Sarvana, D.~Silver, A.~Chen, L.~Wang, L.~Maggiore, O.~Chang, N.~Attaluri, G.~Thornton, C.-C. Chiu, O.~Bunyan, N.~Levine, T.~Chung, E.~Eltyshev, X.~Si, T.~Lillicrap, D.~Brady, V.~Aggarwal, B.~Wu, Y.~Xu, R.~McIlroy, K.~Badola, P.~Sandhu, E.~Moreira, W.~Stokowiec, R.~Hemsley, D.~Li, A.~Tudor, P.~Shyam, E.~Rahimtoroghi, S.~Haykal, P.~Sprechmann, X.~Zhou, D.~Mincu, Y.~Li, R.~Addanki, K.~Krishna, X.~Wu, A.~Frechette, M.~Eyal, A.~Dafoe, D.~Lacey, J.~Whang, T.~Avrahami, Y.~Zhang, E.~Taropa, H.~Lin, D.~Toyama, E.~Rutherford, M.~Sano, H.~Choe, A.~Tomala, C.~Safranek-Shrader, N.~Kassner, M.~Pajarskas, M.~Harvey, S.~Sechrist, M.~Fortunato, C.~Lyu, G.~Elsayed, C.~Kuang, J.~Lottes, E.~Chu, C.~Jia, C.-W. Chen, P.~Humphreys, K.~Baumli,
  C.~Tao, R.~Samuel, C.~N. dos Santos, A.~Andreassen, N.~Rakićević, D.~Grewe, A.~Kumar, S.~Winkler, J.~Caton, A.~Brock, S.~Dalmia, H.~Sheahan, I.~Barr, Y.~Miao, P.~Natsev, J.~Devlin, F.~Behbahani, F.~Prost, Y.~Sun, A.~Myaskovsky, T.~S. Pillai, D.~Hurt, A.~Lazaridou, X.~Xiong, C.~Zheng, F.~Pardo, X.~Li, D.~Horgan, J.~Stanton, M.~Ambar, F.~Xia, A.~Lince, M.~Wang, B.~Mustafa, A.~Webson, H.~Lee, R.~Anil, M.~Wicke, T.~Dozat, A.~Sinha, E.~Piqueras, E.~Dabir, S.~Upadhyay, A.~Boral, L.~A. Hendricks, C.~Fry, J.~Djolonga, Y.~Su, J.~Walker, J.~Labanowski, R.~Huang, V.~Misra, J.~Chen, R.~Skerry-Ryan, A.~Singh, S.~Rijhwani, D.~Yu, A.~Castro-Ros, B.~Changpinyo, R.~Datta, S.~Bagri, A.~M. Hrafnkelsson, M.~Maggioni, D.~Zheng, Y.~Sulsky, S.~Hou, T.~L. Paine, A.~Yang, J.~Riesa, D.~Rogozinska, D.~Marcus, D.~E. Badawy, Q.~Zhang, L.~Wang, H.~Miller, J.~Greer, L.~L. Sjos, A.~Nova, H.~Zen, R.~Chaabouni, M.~Rosca, J.~Jiang, C.~Chen, R.~Liu, T.~Sainath, M.~Krikun, A.~Polozov, J.-B. Lespiau, J.~Newlan, Z.~Cankara, S.~Kwak, Y.~Xu,
  P.~Chen, A.~Coenen, C.~Meyer, K.~Tsihlas, A.~Ma, J.~Gottweis, J.~Xing, C.~Gu, J.~Miao, C.~Frank, Z.~Cankara, S.~Ganapathy, I.~Dasgupta, S.~Hughes-Fitt, H.~Chen, D.~Reid, K.~Rong, H.~Fan, J.~van Amersfoort, V.~Zhuang, A.~Cohen, S.~S. Gu, A.~Mohananey, A.~Ilic, T.~Tobin, J.~Wieting, A.~Bortsova, P.~Thacker, E.~Wang, E.~Caveness, J.~Chiu, E.~Sezener, A.~Kaskasoli, S.~Baker, K.~Millican, M.~Elhawaty, K.~Aisopos, C.~Lebsack, N.~Byrd, H.~Dai, W.~Jia, M.~Wiethoff, E.~Davoodi, A.~Weston, L.~Yagati, A.~Ahuja, I.~Gao, G.~Pundak, S.~Zhang, M.~Azzam, K.~C. Sim, S.~Caelles, J.~Keeling, A.~Sharma, A.~Swing, Y.~Li, C.~Liu, C.~G. Bostock, Y.~Bansal, Z.~Nado, A.~Anand, J.~Lipschultz, A.~Karmarkar, L.~Proleev, A.~Ittycheriah, S.~H. Yeganeh, G.~Polovets, A.~Faust, J.~Sun, A.~Rrustemi, P.~Li, R.~Shivanna, J.~Liu, C.~Welty, F.~Lebron, A.~Baddepudi, S.~Krause, E.~Parisotto, R.~Soricut, Z.~Xu, D.~Bloxwich, M.~Johnson, B.~Neyshabur, J.~Mao-Jones, R.~Wang, V.~Ramasesh, Z.~Abbas, A.~Guez, C.~Segal, D.~D. Nguyen, J.~Svensson, L.~Hou,
  S.~York, K.~Milan, S.~Bridgers, W.~Gworek, M.~Tagliasacchi, J.~Lee-Thorp, M.~Chang, A.~Guseynov, A.~J. Hartman, M.~Kwong, R.~Zhao, S.~Kashem, E.~Cole, A.~Miech, R.~Tanburn, M.~Phuong, F.~Pavetic, S.~Cevey, R.~Comanescu, R.~Ives, S.~Yang, C.~Du, B.~Li, Z.~Zhang, M.~Iinuma, C.~H. Hu, A.~Roy, S.~Bijwadia, Z.~Zhu, D.~Martins, R.~Saputro, A.~Gergely, S.~Zheng, D.~Jia, I.~Antonoglou, A.~Sadovsky, S.~Gu, Y.~Bi, A.~Andreev, S.~Samangooei, M.~Khan, T.~Kocisky, A.~Filos, C.~Kumar, C.~Bishop, A.~Yu, S.~Hodkinson, S.~Mittal, P.~Shah, A.~Moufarek, Y.~Cheng, A.~Bloniarz, J.~Lee, P.~Pejman, P.~Michel, S.~Spencer, V.~Feinberg, X.~Xiong, N.~Savinov, C.~Smith, S.~Shakeri, D.~Tran, M.~Chesus, B.~Bohnet, G.~Tucker, T.~von Glehn, C.~Muir, Y.~Mao, H.~Kazawa, A.~Slone, K.~Soparkar, D.~Shrivastava, J.~Cobon-Kerr, M.~Sharman, J.~Pavagadhi, C.~Araya, K.~Misiunas, N.~Ghelani, M.~Laskin, D.~Barker, Q.~Li, A.~Briukhov, N.~Houlsby, M.~Glaese, B.~Lakshminarayanan, N.~Schucher, Y.~Tang, E.~Collins, H.~Lim, F.~Feng, A.~Recasens, G.~Lai,
  A.~Magni, N.~D. Cao, A.~Siddhant, Z.~Ashwood, J.~Orbay, M.~Dehghani, J.~Brennan, Y.~He, K.~Xu, Y.~Gao, C.~Saroufim, J.~Molloy, X.~Wu, S.~Arnold, S.~Chang, J.~Schrittwieser, E.~Buchatskaya, S.~Radpour, M.~Polacek, S.~Giordano, A.~Bapna, S.~Tokumine, V.~Hellendoorn, T.~Sottiaux, S.~Cogan, A.~Severyn, M.~Saleh, S.~Thakoor, L.~Shefey, S.~Qiao, M.~Gaba, S.~yiin Chang, C.~Swanson, B.~Zhang, B.~Lee, P.~K. Rubenstein, G.~Song, T.~Kwiatkowski, A.~Koop, A.~Kannan, D.~Kao, P.~Schuh, A.~Stjerngren, G.~Ghiasi, G.~Gibson, L.~Vilnis, Y.~Yuan, F.~T. Ferreira, A.~Kamath, T.~Klimenko, K.~Franko, K.~Xiao, I.~Bhattacharya, M.~Patel, R.~Wang, A.~Morris, R.~Strudel, V.~Sharma, P.~Choy, S.~H. Hashemi, J.~Landon, M.~Finkelstein, P.~Jhakra, J.~Frye, M.~Barnes, M.~Mauger, D.~Daun, K.~Baatarsukh, M.~Tung, W.~Farhan, H.~Michalewski, F.~Viola, F.~de~Chaumont~Quitry, C.~L. Lan, T.~Hudson, Q.~Wang, F.~Fischer, I.~Zheng, E.~White, A.~Dragan, J.~baptiste Alayrac, E.~Ni, A.~Pritzel, A.~Iwanicki, M.~Isard, A.~Bulanova, L.~Zilka, E.~Dyer,
  D.~Sachan, S.~Srinivasan, H.~Muckenhirn, H.~Cai, A.~Mandhane, M.~Tariq, J.~W. Rae, G.~Wang, K.~Ayoub, N.~FitzGerald, Y.~Zhao, W.~Han, C.~Alberti, D.~Garrette, K.~Krishnakumar, M.~Gimenez, A.~Levskaya, D.~Sohn, J.~Matak, I.~Iturrate, M.~B. Chang, J.~Xiang, Y.~Cao, N.~Ranka, G.~Brown, A.~Hutter, V.~Mirrokni, N.~Chen, K.~Yao, Z.~Egyed, F.~Galilee, T.~Liechty, P.~Kallakuri, E.~Palmer, S.~Ghemawat, J.~Liu, D.~Tao, C.~Thornton, T.~Green, M.~Jasarevic, S.~Lin, V.~Cotruta, Y.-X. Tan, N.~Fiedel, H.~Yu, E.~Chi, A.~Neitz, J.~Heitkaemper, A.~Sinha, D.~Zhou, Y.~Sun, C.~Kaed, B.~Hulse, S.~Mishra, M.~Georgaki, S.~Kudugunta, C.~Farabet, I.~Shafran, D.~Vlasic, A.~Tsitsulin, R.~Ananthanarayanan, A.~Carin, G.~Su, P.~Sun, S.~V, G.~Carvajal, J.~Broder, I.~Comsa, A.~Repina, W.~Wong, W.~W. Chen, P.~Hawkins, E.~Filonov, L.~Loher, C.~Hirnschall, W.~Wang, J.~Ye, A.~Burns, H.~Cate, D.~G. Wright, F.~Piccinini, L.~Zhang, C.-C. Lin, I.~Gog, Y.~Kulizhskaya, A.~Sreevatsa, S.~Song, L.~C. Cobo, A.~Iyer, C.~Tekur, G.~Garrido, Z.~Xiao,
  R.~Kemp, H.~S. Zheng, H.~Li, A.~Agarwal, C.~Ngani, K.~Goshvadi, R.~Santamaria-Fernandez, W.~Fica, X.~Chen, C.~Gorgolewski, S.~Sun, R.~Garg, X.~Ye, S.~M.~A. Eslami, N.~Hua, J.~Simon, P.~Joshi, Y.~Kim, I.~Tenney, S.~Potluri, L.~N. Thiet, Q.~Yuan, F.~Luisier, A.~Chronopoulou, S.~Scellato, P.~Srinivasan, M.~Chen, V.~Koverkathu, V.~Dalibard, Y.~Xu, B.~Saeta, K.~Anderson, T.~Sellam, N.~Fernando, F.~Huot, J.~Jung, M.~Varadarajan, M.~Quinn, A.~Raul, M.~Le, R.~Habalov, J.~Clark, K.~Jalan, K.~Bullard, A.~Singhal, T.~Luong, B.~Wang, S.~Rajayogam, J.~Eisenschlos, J.~Jia, D.~Finchelstein, A.~Yakubovich, D.~Balle, M.~Fink, S.~Agarwal, J.~Li, D.~Dvijotham, S.~Pal, K.~Kang, J.~Konzelmann, J.~Beattie, O.~Dousse, D.~Wu, R.~Crocker, C.~Elkind, S.~R. Jonnalagadda, J.~Lee, D.~Holtmann-Rice, K.~Kallarackal, R.~Liu, D.~Vnukov, N.~Vats, L.~Invernizzi, M.~Jafari, H.~Zhou, L.~Taylor, J.~Prendki, M.~Wu, T.~Eccles, T.~Liu, K.~Kopparapu, F.~Beaufays, C.~Angermueller, A.~Marzoca, S.~Sarcar, H.~Dib, J.~Stanway, F.~Perbet, N.~Trdin,
  R.~Sterneck, A.~Khorlin, D.~Li, X.~Wu, S.~Goenka, D.~Madras, S.~Goldshtein, W.~Gierke, T.~Zhou, Y.~Liu, Y.~Liang, A.~White, Y.~Li, S.~Singh, S.~Bahargam, M.~Epstein, S.~Basu, L.~Lao, A.~Ozturel, C.~Crous, A.~Zhai, H.~Lu, Z.~Tung, N.~Gaur, A.~Walton, L.~Dixon, M.~Zhang, A.~Globerson, G.~Uy, A.~Bolt, O.~Wiles, M.~Nasr, I.~Shumailov, M.~Selvi, F.~Piccinno, R.~Aguilar, S.~McCarthy, M.~Khalman, M.~Shukla, V.~Galic, J.~Carpenter, K.~Villela, H.~Zhang, H.~Richardson, J.~Martens, M.~Bosnjak, S.~R. Belle, J.~Seibert, M.~Alnahlawi, B.~McWilliams, S.~Singh, A.~Louis, W.~Ding, D.~Popovici, L.~Simicich, L.~Knight, P.~Mehta, N.~Gupta, C.~Shi, S.~Fatehi, J.~Mitrovic, A.~Grills, J.~Pagadora, D.~Petrova, D.~Eisenbud, Z.~Zhang, D.~Yates, B.~Mittal, N.~Tripuraneni, Y.~Assael, T.~Brovelli, P.~Jain, M.~Velimirovic, C.~Akbulut, J.~Mu, W.~Macherey, R.~Kumar, J.~Xu, H.~Qureshi, G.~Comanici, J.~Wiesner, Z.~Gong, A.~Ruddock, M.~Bauer, N.~Felt, A.~GP, A.~Arnab, D.~Zelle, J.~Rothfuss, B.~Rosgen, A.~Shenoy, B.~Seybold, X.~Li,
  J.~Mudigonda, G.~Erdogan, J.~Xia, J.~Simsa, A.~Michi, Y.~Yao, C.~Yew, S.~Kan, I.~Caswell, C.~Radebaugh, A.~Elisseeff, P.~Valenzuela, K.~McKinney, K.~Paterson, A.~Cui, E.~Latorre-Chimoto, S.~Kim, W.~Zeng, K.~Durden, P.~Ponnapalli, T.~Sosea, C.~A. Choquette-Choo, J.~Manyika, B.~Robenek, H.~Vashisht, S.~Pereira, H.~Lam, M.~Velic, D.~Owusu-Afriyie, K.~Lee, T.~Bolukbasi, A.~Parrish, S.~Lu, J.~Park, B.~Venkatraman, A.~Talbert, L.~Rosique, Y.~Cheng, A.~Sozanschi, A.~Paszke, P.~Kumar, J.~Austin, L.~Li, K.~Salama, W.~Kim, N.~Dukkipati, A.~Baryshnikov, C.~Kaplanis, X.~Sheng, Y.~Chervonyi, C.~Unlu, D.~de~Las~Casas, H.~Askham, K.~Tunyasuvunakool, F.~Gimeno, S.~Poder, C.~Kwak, M.~Miecnikowski, V.~Mirrokni, A.~Dimitriev, A.~Parisi, D.~Liu, T.~Tsai, T.~Shevlane, C.~Kouridi, D.~Garmon, A.~Goedeckemeyer, A.~R. Brown, A.~Vijayakumar, A.~Elqursh, S.~Jazayeri, J.~Huang, S.~M. Carthy, J.~Hoover, L.~Kim, S.~Kumar, W.~Chen, C.~Biles, G.~Bingham, E.~Rosen, L.~Wang, Q.~Tan, D.~Engel, F.~Pongetti, D.~de~Cesare, D.~Hwang, L.~Yu,
  J.~Pullman, S.~Narayanan, K.~Levin, S.~Gopal, M.~Li, A.~Aharoni, T.~Trinh, J.~Lo, N.~Casagrande, R.~Vij, L.~Matthey, B.~Ramadhana, A.~Matthews, C.~Carey, M.~Johnson, K.~Goranova, R.~Shah, S.~Ashraf, K.~Dasgupta, R.~Larsen, Y.~Wang, M.~R. Vuyyuru, C.~Jiang, J.~Ijazi, K.~Osawa, C.~Smith, R.~S. Boppana, T.~Bilal, Y.~Koizumi, Y.~Xu, Y.~Altun, N.~Shabat, B.~Bariach, A.~Korchemniy, K.~Choo, O.~Ronneberger, C.~Iwuanyanwu, S.~Zhao, D.~Soergel, C.-J. Hsieh, I.~Cai, S.~Iqbal, M.~Sundermeyer, Z.~Chen, E.~Bursztein, C.~Malaviya, F.~Biadsy, P.~Shroff, I.~Dhillon, T.~Latkar, C.~Dyer, H.~Forbes, M.~Nicosia, V.~Nikolaev, S.~Greene, M.~Georgiev, P.~Wang, N.~Martin, H.~Sedghi, J.~Zhang, P.~Banzal, D.~Fritz, V.~Rao, X.~Wang, J.~Zhang, V.~Patraucean, D.~Du, I.~Mordatch, I.~Jurin, L.~Liu, A.~Dubey, A.~Mohan, J.~Nowakowski, V.-D. Ion, N.~Wei, R.~Tojo, M.~A. Raad, D.~A. Hudson, V.~Keshava, S.~Agrawal, K.~Ramirez, Z.~Wu, H.~Nguyen, J.~Liu, M.~Sewak, B.~Petrini, D.~Choi, I.~Philips, Z.~Wang, I.~Bica, A.~Garg, J.~Wilkiewicz,
  P.~Agrawal, X.~Li, D.~Guo, E.~Xue, N.~Shaik, A.~Leach, S.~M. Khan, J.~Wiesinger, S.~Jerome, A.~Chakladar, A.~W. Wang, T.~Ornduff, F.~Abu, A.~Ghaffarkhah, M.~Wainwright, M.~Cortes, F.~Liu, J.~Maynez, S.~Petrov, Y.~Wu, D.~Hassabis, K.~Kavukcuoglu, J.~Dean, and O.~Vinyals, ``Gemini 1.5: Unlocking multimodal understanding across millions of tokens of context,'' 2024.

\bibitem{LIWC-22}
R.~Boyd, A.~Ashokkumar, S.~Seraj, and J.~Pennebaker, ``The development and psychometric properties of liwc-22,'' 02 2022.

\end{thebibliography}
\bibliographystyle{ieeetr}
\end{document}